\documentclass[letterpaper, 10 pt, conference]{ieeeconf}  

\IEEEoverridecommandlockouts                              
\usepackage{graphicx}
\usepackage{float}
\usepackage{gensymb}
\usepackage{amsmath}
\usepackage{hyperref}
\usepackage{booktabs}
\usepackage{tabularx}
\usepackage{multirow}

\hypersetup{
    colorlinks=true,
}
\usepackage{algorithmic}

\usepackage[ruled,vlined]{algorithm2e}
\SetKw{Continue}{continue}
\SetKw{Break}{break}
\LinesNumbered

\usepackage{cite}

\title{\LARGE \bf
Informative Sensor Planning for a Single-Axis Gimbaled Camera on a Fixed-Wing UAV 
}

\author{Aditya Parandekar$^{1}$, Brady Moon$^{2}$, Nayana Suvarna$^{2}$, and Sebastian Scherer$^{2}$
\thanks{This work is supported by the Office of Naval Research (Grant N00014-21-1-2110). This material is based upon work supported by the National Science Foundation Graduate Research Fellowship under Grant No. DGE1745016.}
\thanks{$^{1}$ A. Parandekar is with Birla Institute of Technology and Science, Pilani – Goa Campus, Sancoale, Goa 403726, India. {\tt\footnotesize  f20201879@goa.bits-pilani.ac.in}}%
\thanks{$^{2}$ B. Moon, N. Suvarna, and S. Scherer are with the Robotics Institute, School of Computer Science at Carnegie Mellon University, Pittsburgh, PA, USA
        {\tt\footnotesize \{bradym, nsuvarna, basti\}@andrew.cmu.edu}}
}

\begin{document}

\maketitle
\thispagestyle{empty}
\pagestyle{empty}

\begin{abstract}

Uncrewed Aerial Vehicles (UAVs) are a leading choice of platforms for a variety of information-gathering applications. Sensor planning can enhance the efficiency and success of these types of missions when coupled with a higher-level informative path-planning algorithm. This paper aims to address these data acquisition challenges by developing an informative non-myopic sensor planning framework for a single-axis gimbal coupled with an informative path planner to maximize information gain over a prior information map. This is done by finding reduced sensor sweep bounds over a planning horizon such that regions of higher confidence are prioritized. This novel sensor planning framework is evaluated against a predefined sensor sweep and no sensor planning baselines as well as validated in two simulation environments. In our results, we observe an improvement in the performance by \textit{21.88\%} and \textit{13.34\%} for the no sensor planning and predefined sensor sweep baselines respectively. 

\end{abstract}

\section{Introduction}

Uncrewed Aerial Vehicles (UAVs) are currently employed for data gathering tasks such as surface monitoring \cite{Bircher2016, COLOMINA201479}, searching for and tracking objects \cite{Nguyen2020}, environmental sciences \cite{env, Popovi2020, patrikar2020wind}, and reconnaissance missions \cite{LIU2022108653, Chen2022, wang2019reconnaissance}. Multi-rotor UAVs are often used in the above data-gathering tasks due to being fairly simple to operate, not requiring a large space to take off, and being fairly portable. However, some limitations include low flight speeds and short battery life. Fixed-wing UAVs are an alternative that offers longer flight times and higher flight speeds. Vertical Take-Off and Landing (VTOL) fixed-wing UAVs are also an option that brings together the benefits of both flight profiles, including being able to take off without a runway.

Cameras, such as Electro-Optical and Infrared Sensors (EO/IR) sensors, are a common payload for data gathering and when paired with a fixed-wing UAV can efficiently cover a large space. However, because these cameras have a restricted field of view in the shape of a frustum rather than being isotropic, the planning problem can become much more complex, especially when there are dynamic constraints between the sensor and vehicle. By mounting the camera on a three-axis gimbal, the rotation of the camera is decoupled from the rotation of the UAV and can then be utilized to efficiently cover a large space. However, these payloads are more expensive, heavy, and large which can limit the scale of adaptability when deploying them on teams of UAVs and limit the minimum size of the aircraft. The extra moving parts also introduce more modes for failure and must be robust to high jerks caused by some launch sequences. 

\begin{figure}
    \centering
    \includegraphics[width=1\linewidth]{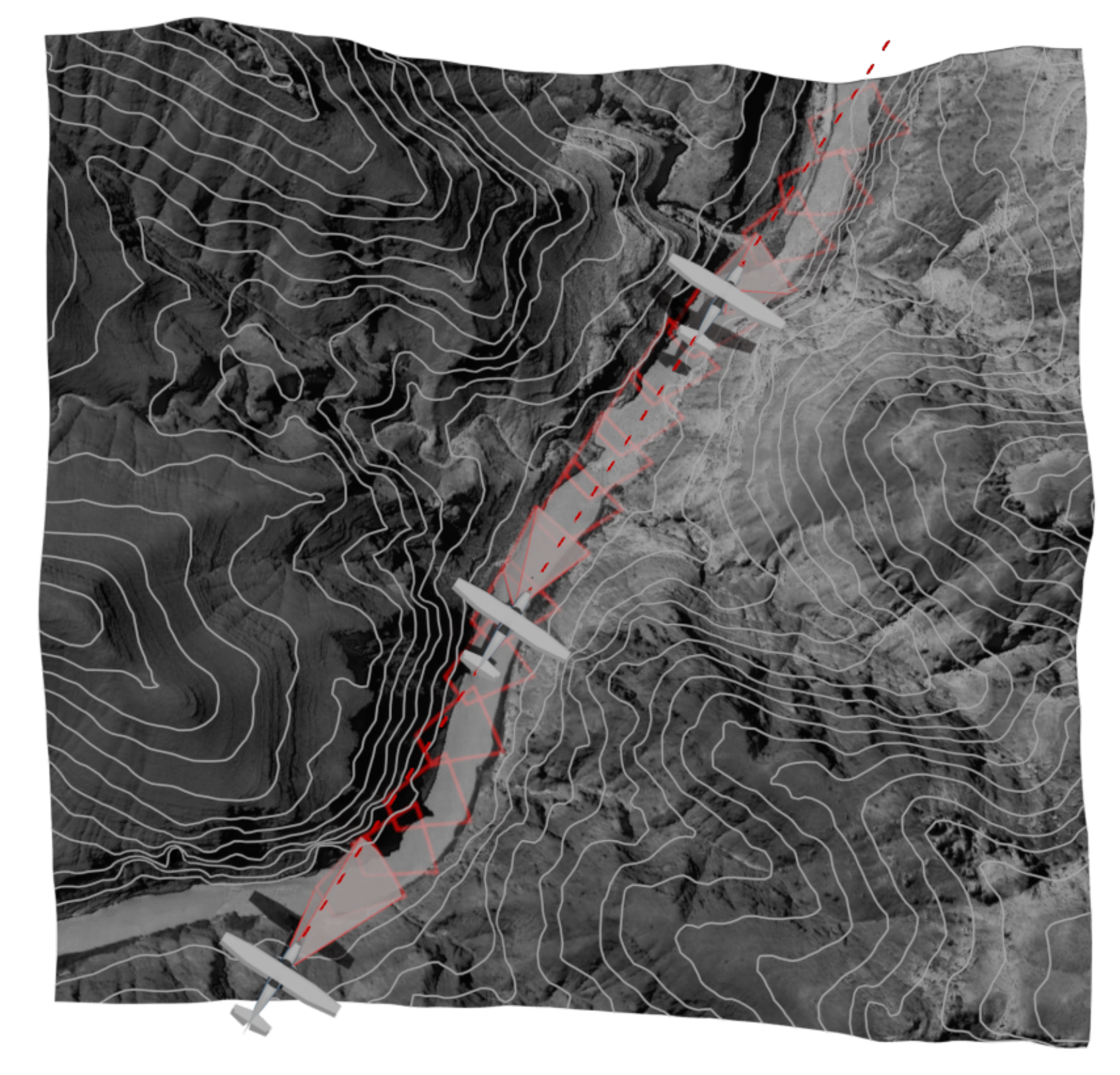}
    \caption{An example scenario of a UAV sweeping a gimbaled camera to search for a missing hiker. As the UAV follows its straight line path over the mountain, the sensor planner prioritizes sweeping over the river.}
    \label{fig:needforsensorplanner}
\end{figure}

A typical solution is to use a fixed-mount camera on the UAV which reduces the dimensionality of the planning problem while also removing the risk of a hardware malfunction on a gimbal, minimizing costs, and reducing the size and weight of the payload. 
Fixed-mount cameras are typically mounted facing downward or forward. These configurations greatly reduce the amount of area viewed, especially in the downward-facing configuration, thus reducing the efficiency of the data gathering. There exists a need for a lower-cost solution that combines the benefits of a gimbaled camera while reducing its limitations. 

One way to do this is by mounting the camera on a single-axis gimbal which allows the camera to sweep in one dimension, greatly increasing the coverage area while still having a fairly small profile, low cost, and high robustness to hardware failures. Additionally, adding only one extra degree of freedom keeps the dimension of the problem low while still producing large benefits. A forward facing camera sweeping side to side allows not only increases in the width of the sensor swaths but also allows for the focusing of the camera angle on areas of interest.

Planning paths that maximize the amount of data, or information, gathered by the camera on the UAV is often formulated as an Informative Path Planning (IPP) problem. This problem formulation maximizes an information objective function while not violating a budget constraint, such as flight time or battery power. 

There have been multiple works done in the past to solve the IPP problem. A few methods providing sub-optimal solutions were proposed in the past. For example, greedy, myopic approaches like \cite{articlegreedy} maximize information gain according to the current state and don't reason over the future states. To counter this complete greediness, algorithms such as \cite{recursivegreedy} were employed to discard sub-optimal solutions by using a recursive greedy strategy to converge on a near-optimal solution in polynomial time. These approaches still failed in finding optimal solutions to IPP tasks. Hence, the need for research on more non-myopic approaches became evident. Methods described in \cite{nonmyopic1}, \cite{nonmyopic2} use a non-myopic approach to solve this IPP problem. One more non-myopic approach is described in \cite{nonmyopic3} for a UAV glider. After comparing the above-mentioned works it is clear that non myopic non greedy approaches outperform myopic greedy approaches for solving these IPP problems. 

In some of the earlier works in the field of sensor planning for data gathering, the sensors are either restricted by their field of view (FOV) or by some other environmental reason. These sensors may be fixed to the UAV body or may move following a fixed predefined trajectory to improve data collection, as described in \cite{fixedsensormovement}. 

Our work presents a hierarchical planning approach that decouples sensor and vehicle planning. This keeps the dimensionality of the vehicle planning problem low while still augmenting the planner to take into account the extra coverage provided by the gimbaled camera. The sensor planner can then run at a higher rate and adapt quickly to current measurements to efficiently gather data.

This sensor planner employs a grid cell evaluation methodology within the search space to identify high-information points. These points determine the sweep bounds for the gimbaled camera's yaw direction, facilitating efficient viewing of these high-information grid cells, and thereby enhancing search mission efficiency.
Hence the key contributions of this work are as follows:
\begin{itemize}
\item An informative non-myopic sensor planning framework to estimate sweep bounds for a single-axis gimbaled camera for increasing the efficiency of a higher-level IPP planner
\item Global IPP planner modifications to account for a sweeping sensor planner
\item Validation of our approach against a pre-defined and no sensor sweep approach in a simulated end-to-end system
\end{itemize}

This paper is organized as follows: Section \ref{sec:PD} provides our problem formulation. Section \ref{sec:Approach} outlines our approach for the gimbal planner and how we integrate it into a global path planner. Section \ref{sec:results} presents our results, and we give our conclusion and future work in Section \ref{sec:conclusion}.

\section{Problem Formulation}\label{sec:PD}

We formulate the problem as finding the sweep bounds, $\psi_1$ and $\psi_2$, for a single-axis gimbal over a sensor planning horizon $h$ while following the planned path provided by the global IPP algorithm. Let the camera state $\psi_c$ be bounded by the max camera angle $\psi_{max}$. Let $\textit{T}$ represent the UAV trajectory. The state vector of the UAV is defined as
\begin{equation*}
\mathbf{x} = [x \quad y \quad z \quad \psi].
\end{equation*}
The camera angle is defined about the z-axis of our UAV state and has a constant pitch angle of $\theta_c$.

\section{Proposed Approach} \label{sec:Approach}

\subsection{The Sensor Planning Algorithm}
\begin{figure}
    \centering
    \includegraphics[width=1\linewidth]{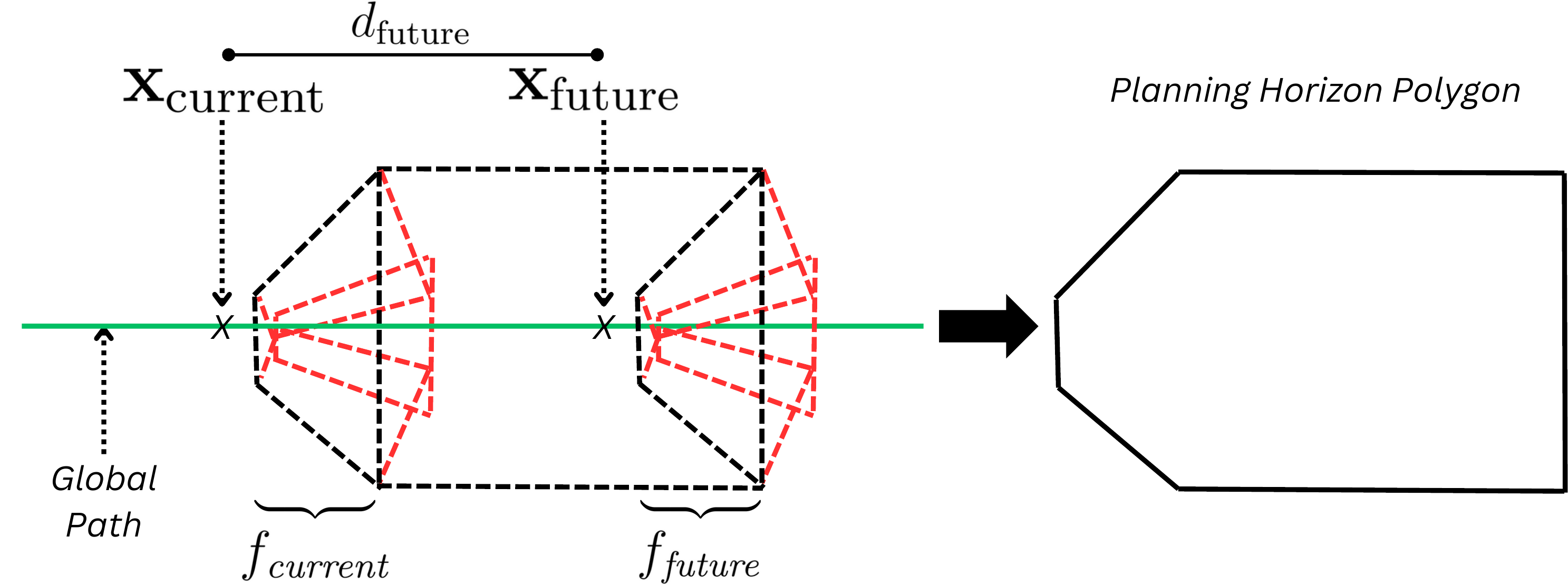}
    \caption{A visualization of how $f_{\text{\textit{current}}}$ and $f_{\text{\textit{current}}}$ are created and combined to create a planner horizon polygon.}
    \label{fig:PlanningHorizon}
\end{figure}

To dynamically update the sensor sweep bounds when the UAV is following a straight-line trajectory, a systematic evaluation of the search space is used to identify locations of high information. 
We define $f$ as the combined projected camera footprint as the gimbal sweeps from a maximum to minimum yaw angle at a UAV state $\mathbf{x}$ and then connecting the outer edges to form a trapezoid, as seen in Fig. \ref{fig:PlanningHorizon}. The red dashed lines show individual sensor footprints which are connected to form the trapezoid with black dashed lines. 
We define a sensor planning horizon polygon using $f_{\text{\textit{current}}}$ and $f_{\text{\textit{future}}}$ trapezoids, which
represents the regions of the search space that the camera views when traveling from $\mathbf{x_{\text{current}}}$ to $\mathbf{x_{\text{future}}}$.
This process is shown in Fig. \ref{fig:PlanningHorizon}. 

\begin{algorithm}[t]
\caption{Calculate Future Position}\label{alg:expected-position}
\SetInd{0.4em}{0.8em}
\SetKwInOut{Input}{Input}
\SetKwInOut{Output}{Output}
\Input{\textit{plan, wp, speed}}
\Output{Future state $\mathbf{x_{\text{future}}}$}
\BlankLine
    $d_{future} \gets speed \times (t_{max\_sweep} + t_{{future}})$\;
    $d_{covered} \gets 0$, $i \gets wp$\;
    \While{$i < \text{len}(plan)$ and $d_{covered} < d_{future}$}{
        $d_{covered} \gets d_{covered} + \text{dist}(plan[i], pos_{agent})$\;
        $i \gets i + 1$\;
    }
    \If{$i < \text{len}(plan)$}{
        $d_{rem} \gets d_{total} - d_{covered}$\;
        $d_{seg} \gets \text{dist}(plan[i], plan[i-1])$\;
        $ratio \gets d_{rem} / d_{seg}$\;
        \Return $\text{interp}(plan[i-1], plan[i], ratio)$\;
    }
    \Else{
        \Return \textit{$\text{current\_pos}$\;}
    }
\end{algorithm}

To find $f_{\text{\textit{future}}}$ for the planning horizon, $\mathbf{x_{\text{future}}}$ is estimated using Alg. \ref{alg:expected-position}. 
It predicts a UAV's future position by calculating the distance from the UAV's current state to the future state $d_{\text{future}} = \text{speed} \times (t_{\text{max\_sweep}} + t_{\text{future}})$ it can travel within the maximum sweep time $t_{\text{max\_sweep}}$ (to ensure we never miss a high information region) and a specified time $t_{\text{future}}$ for which are reasoning over to find the future state of the UAV on its global path. Starting at the current state $\mathbf{x_{\text{current}}}$ of the UAV, we iteratively add up the distance traveled to each waypoint \textit{wp} until reaching a distance greater than or equal to $d_{\text{future}}$. We then interpolate the future UAV position using the ratio of remaining segment distance, hence returning the future state, $\mathbf{x_{\text{future}}}$ of the UAV. If the end of the plan is reached while iterating up to $d_{\text{future}}$, then we just return the final state in the plan. 

After estimating $\mathbf{x_{\text{future}}}$ of the UAV on its global planned trajectory, the $f_{\text{\textit{future}}}$ is formed using the maximum limit of the sweep bounds at this newly found $\mathbf{x_{\text{future}}}$ for the UAV. This estimated $f_{\text{\textit{future}}}$ is combined with the initially defined $f_{\text{\textit{current}}}$ to complete the planning horizon polygon (shown in Fig. \ref{fig:PlanningHorizon}). This forms our planning horizon or region which we evaluate to find the boundary high information grid cells to finally estimate the optimal sweep bounds for the gimbaled camera based on (\ref{alg:alg1}). 

\begin{algorithm}[t]
\caption{Find Boundary High-Information Cell}
\label{alg:alg1}
\SetInd{0.4em}{0.8em}
\SetKwInOut{Input}{Input}
\SetKwInOut{Output}{Output}
\SetKwFunction{FCalcY}{CalcLayer}
\SetKwFunction{FFindVerts}{FindEndpoints}
\SetKwFunction{FFindHighInfo}{FindHighInfoCell}
\SetKwProg{Fn}{Function}{:}{}
\Input{$f_{\text{\textit{current}}}, f_{\text{\textit{future}}}, n_{\text{layers}}$, $is\_upper$} 
\Output{High info cells on boundaries}

\BlankLine
$layerHt \leftarrow$ vertical iteration param\;
$layer \leftarrow 0$\;
\While{$layer < n_{\text{layers}}$}{
    $curY \leftarrow$ \FCalcY{$layer$, $layerHt$, $is\_upper$}\;
    $\textbf{v} \leftarrow$ \FFindVerts{$f_{\text{\textit{future}}}$, $f_{\text{\textit{current}}}$, $curY$}\;
    $cells \leftarrow$ cells along line from $\textbf{v}$\;
    $highInfoCell \leftarrow$ \FFindHighInfo{$cells$}\;
    \If{$highInfoCell$ \text{is not empty}}{
        \textbf{break}\;
    }
    $layer \leftarrow layer + 1$\;
}
\Return{$highInfoCell$}\;
\end{algorithm}

Alg. \ref{alg:alg1} describes the process followed to estimate these boundary-high information grid cells. Instead of iterating over every grid cell in the planning horizon, we follow a systematic layer-by-layer selective grid evaluation to estimate the boundary high information grid cells. By starting from the outsides working inward, once a cell above a specified information threshold is found, we don't have to continue evaluating. This is done both on the upper and lower sides of the planning horizon polygon described in Fig. \ref{fig:PlanningHorizon}. We define a vertical separation between each layer, defined as $layerHt$, which is found by dividing the planning horizon polygon height by $n_{layers}$. We then use this vertical separation to calculate the current layer orthogonal distance $curY$ from the UAV path (seen as the green line in Fig.~\ref{fig:PlanningHorizon}). When calculating $curY$ for the upper half of the planning horizon polygon, the boolean $is\_upper$ is set to true, and false for the bottom. Then we use $curY$ to find the endpoints of that particular layer. These endpoints $\mathbf{v}$ are passed to a modified version of Bresenham's line algorithm described in \cite{articlebresenham} to find the cells in that particular layer. This list of cells is evaluated till a high information grid cell is found using Alg.~\ref{alg:alg2}. If found, the algorithm returns the coordinate of the high information grid cell which we then use to decide our sensor sweep bounds. 
Because the algorithm iterates through the cells in the planning horizon polygon from outside to inside, 
it always ensures that the high-information cells nearer to the center of the polygon are also viewed and no high-information cells are missed. 

For the portions of the path where the UAV is banking for a turn, the gimbaled camera performs a full sweep. We identify where the UAV will be turning by utilizing the waypoint information \textit{wp} from the global IPP planner and evaluate sequential waypoints to determine if the UAV is turning or not over the duration of our planning horizon. If the UAV is expected to turn during the course of the planning horizon, the sweep bounds are then set to the maximum bounds.
Future work could implement optimal sweep bounds over these turns or simply find a fixed sweep bounds during turns that avoids having the sensor point toward the sky where no information is gained. 

\begin{figure*}[th]
    \centering
    \includegraphics[trim={0cm 0cm 0cm 0cm}, width=7in]{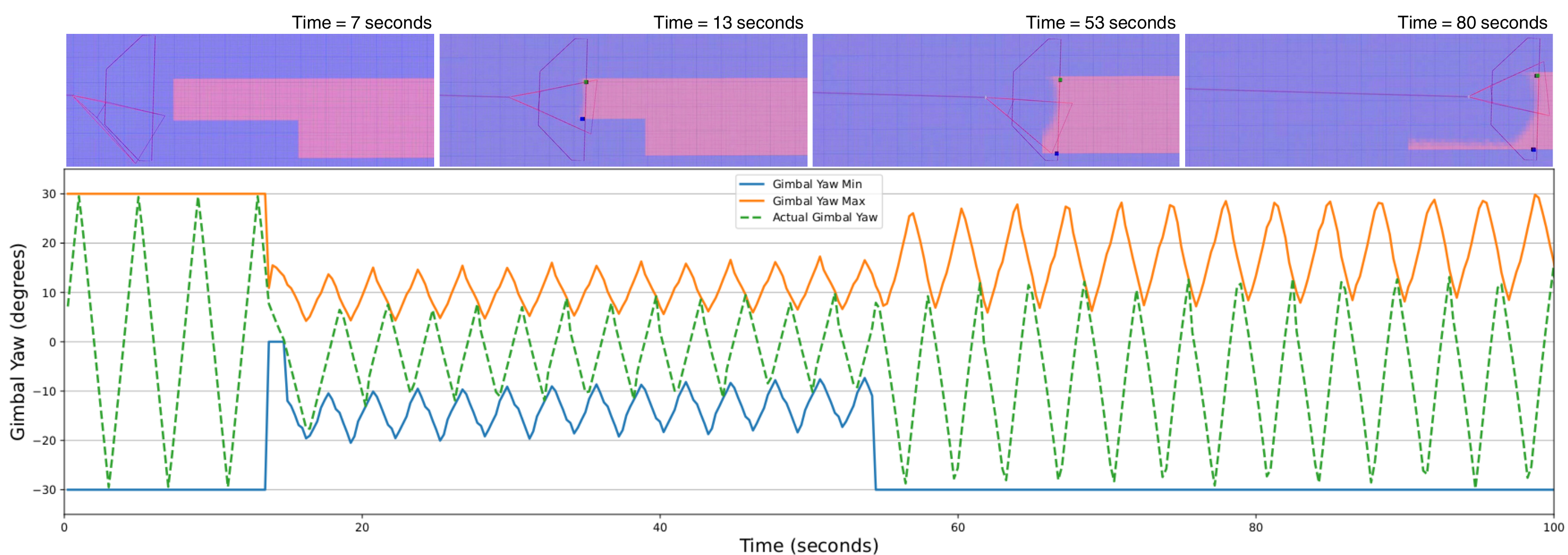}
    \caption{An example scenario showing the belief space as well as the output of our sensor planner. You can see the boundary high information grid cells marked with green and blue squares in the top row of images. The actual gimbal yaw is plotted along with the gimbal bounds to show how the gimbal sweeps between the bounds. The gimbal focuses on the regions of high information gain and ignores the low-value areas.}
    \label{fig:SensorPlanningAlgo}
\end{figure*} 

Fig.~\ref{fig:SensorPlanningAlgo} shows an example scenario of deciding gimbal yaw sweep bounds by finding boundary-high information grid cells. The gimbaled camera initially, when there are no high information grid cells inside the planning horizon polygon, sweeps till its maximum limit i.e., $-30^\circ$ to $30^\circ$. At time $t=13$ seconds, the planning horizon polygon encounters high information grid cells (pink-colored cells). It then uses Alg. \ref{alg:alg1} to find boundary-high information grid cells on both halves of the polygon (shown by the green and blue cubes), calculating the gimbal yaw bounds to view these identified high information cells. As observed in the second image of Fig.~\ref{fig:SensorPlanningAlgo}, the sweep bounds change from max to min sweep to a constrained sweep pattern estimated after selectively evaluating the search space. It then changes its sweep pattern downwards to view the newly encountered high information region at time $t=53$ seconds. The sweep pattern adjusts according to the estimated boundary high information cells while following a global planned trajectory for the UAV. 

\subsection{Global IPP Planner Integration}

This sensor planning algorithm is integrated with TIGRIS \cite{moon2022tigris}, an informed sampling-based IPP algorithm that finds paths that maximize information gain. TIGRIS performs well at finding planning paths in high dimensions and larger search spaces while staying within a path constrained by a budget $B$. 
This IPP problem is defined as
\begin{equation*}
    \label{tigirseqn}
    T^* = \underset{T \in \mathcal{T}}{\text{argmax}} \, I(T) \quad \text{s.t.} \quad C(T) \leq B.
\end{equation*}
To combine the global planner with our gimbal planner, we choose to take a cascaded approach where we first create global plans with TIGRIS and then create gimbal plans as defined in the previous section. The main modification to TIGRIS is to artificially widen the FOV for the sensor. By increasing the width of the sensor, we ensure that it now takes into account the entire maximum to the minimum ($\psi_{\text{max}}$ to $\psi_{\text{min}}$) pan of the camera view i.e, the camera projection on the search space. This is because the electro-optical sensor is not static anymore and thus the planner needs to be informed about the accurate width of the camera to plan efficient paths for the search mission.

\subsection{Implementation}
This end-to-end global and gimbal planning method is deployed on a fixed-wing VTOL platform with a forward-facing electro-optical sensor that is pitched downward by $\theta_s$ degrees. The sensor motion is constrained to only rotate side to side and is bounded by a minimum and maximum yaw of $\psi_{min}$ and $\psi_{max}$. 

\begin{algorithm}[t]
\caption{Find High Info Cell}
\label{alg:alg2}
\SetInd{0.4em}{0.8em}
\SetKwInOut{Input}{Input}
\SetKwInOut{Output}{Output}
\SetKwFunction{FInfoReward}{InfoReward} 
\SetKwFunction{FNewBelief}{NewBelief}
\Input{{\textit{cells} from Bresenham's line algorithm}}
\Output{High information cell}
\BlankLine
\ForEach{$cell \in \textit{cells}$}{
    \If{cell\textup{ is inside search map}}{
        $P_{\text{old}} \gets \textit{search\_map}[cell]$\;
        $s_{id} \gets \textit{sensor\_model\_id}[cell]$\;
        $P_{\text{new}} \gets$ \FNewBelief{$P_{\text{old}}$, $s_{id}$}\; 
        $\Delta entropy \gets$ \FInfoReward{$P_{\text{old}}$, $P_{\text{new}}$}\; 
    }
    \If{$\Delta entropy \geq \textit{threshold\_entropy}$}{
        $high\_info\_cell \gets cell$\;
        \Break\;
    }
}
\Return{$high\_info\_cell$}\;
\end{algorithm} 

\subsubsection{The Sensor Model}
The sensor model is used to update the belief of a Bayesian grid search model. This sensor model is created by testing the performance of a perception model on a large quantity of data and evaluating its true positive rate (\textit{tpr}) and false positive rate (\textit{fpr}). The sensors can be modeled to consider different environmental conditions such as sensors' speed and acceleration, and many more. For this use case, we have modeled the sensor in the following manner and defined it as
\begin{equation} \label{3.3}
    f(r) =
    \begin{cases}
         0.9 & \text{if } r \leq \alpha \\
        a\cdot r + b & \text{if } \alpha < r < \beta \\
        0.5 & \text{if } r > \beta \\
    \end{cases}
\end{equation}
wherein $\textit{r}$ is the range of the object of interest from the electro-optical sensor and $\textit{a}$ and $\textit{b}$ are parameters determined through validation of the perception system.
In the equation, $\alpha$ represents the range where the electro-optical sensor performance starts to degrade, and $\beta$ represents the breakpoint for the range $r$ where the sensor measurements \textit{tpr} and \textit{fpr} both converge to $0.5$. Beyond this range, sensor measurements no longer lead to updates in the Bayesian belief space.

\begin{figure}[t]
    \centering
    \includegraphics[trim={0cm 0cm 0cm 0cm},clip,width=0.7\linewidth]{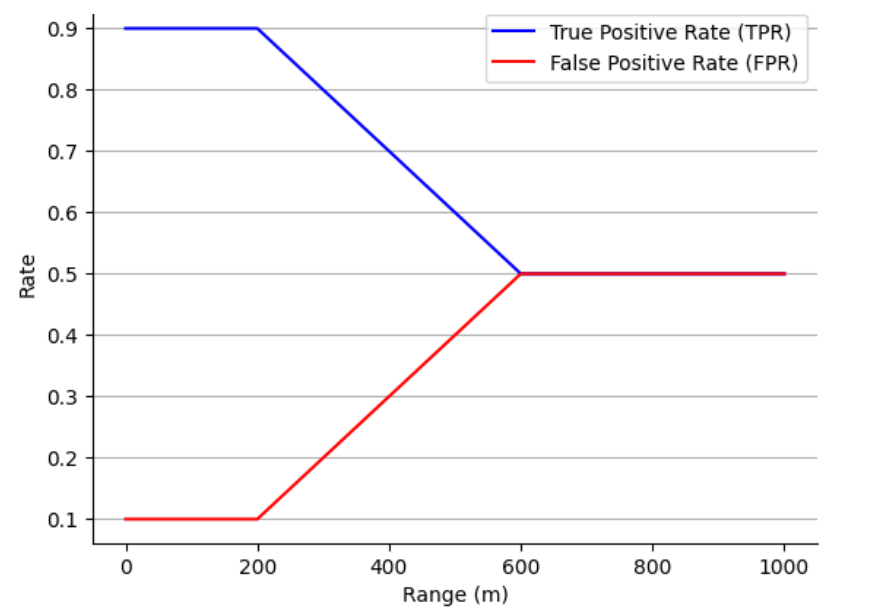}
    \caption{Sensor model used for implementing the proposed sensor planning algorithm based on (\ref{3.3}).}
    \label{fig:sensormodel}
\end{figure}

Hence, we model the sensors \textit{tpr} as \textit{P(Z$|$X)} where its \textit{fpr} is complementary of its \textit{tpr} which is \textit{P(Z$|$-X)}. The performance of the sensor is shown in Fig.~\ref{fig:sensormodel} according to (\ref{3.3}) and its \textit{tpr} and \textit{fpr}. Similarly to model the sensor for incorporating negative measurements, we have defined the \textit{tnr} and \textit{fnr} for the sensor. 

\subsubsection{Information Reward Estimation}
This information gain function is used extensively in planning trajectories via the TIGRIS algorithm and by the proposed sensor planning algorithm. In the algorithm described in Alg.~\ref{alg:alg2} line 9, the entropy is calculated and used in classifying a particular grid cell as \textit{important} or not based on a threshold set for the calculated entropy in Alg.~\ref{alg:alg1}. 

This information entropy is estimated using Shannon entropy \cite{shannonentropy} and is described for our case as
\begin{equation}
    \label{shannonentropy}
    H(X) = -P(X) \log P(X) - P(\neg X) \log P(\neg X).
\end{equation}

We finally want to estimate the entropy reduction due to a particular measurement from the sensor using Bayes update. The entropy reduction for a positive measurement \textit{Z} would be
\begin{equation}
    \label{deltaentropy}
    \Delta H(X|Z) = H(X) - H(X|Z).
\end{equation}

To calculate the entropy mentioned above for estimating the boundary threshold high information grid cells according to Alg. \ref{alg:alg1}, the following method is executed as described below.

We assign values to the new belief $P(X|Z)$ based on the relationship between the set confidence threshold of the old belief and the sensor model where
\begin{equation}
\label{eqn:sensormodelpositive}
P(X | Z) = \frac{{P(Z | X) P(X)}}{{P(Z | X) P(X) + P(Z | \neg X) P(\neg X)}}
\end{equation}

Taking inspiration from \cite{inproceedingslonghorizon}, we define a confidence threshold for estimating the reward function. Suppose the old belief is greater than the confidence threshold. In that case, we assume a positive measurement \textit{Z} for estimating the new belief leveraging the \textit{tpr} and \textit{fpr} of the sensor model, shown in (\ref{eqn:sensormodelpositive}). We follow a similar process for estimating the new belief when the old belief is less than the confidence threshold by leveraging the sensor models \textit{tnr} and \textit{fnr} similar to (\ref{eqn:sensormodelpositive}). Leveraging the Shannon Entropy equation described in (\ref{shannonentropy}), we calculate the entropy for the old and new beliefs. These calculated values are used by  (\ref{deltaentropy}) to estimate the delta entropy of the process. 

These entropy calculations are used in Alg. \ref{alg:alg2} line 6 in the \textit{InfoReward()} function, to find the grid cells with information gain greater than a set threshold. The first grid cell selected above the specified threshold using Alg. \ref{alg:alg1} is then used to find the boundary for the gimbal yaw bounds. By reducing the yaw bounds to not include cells below the threshold, the planner increases the amount of sensor measurements for the cells above the threshold, increasing the total reward and overall reduction in entropy. 

\begin{figure}[t]
    \centering\includegraphics[width=1.0\linewidth]{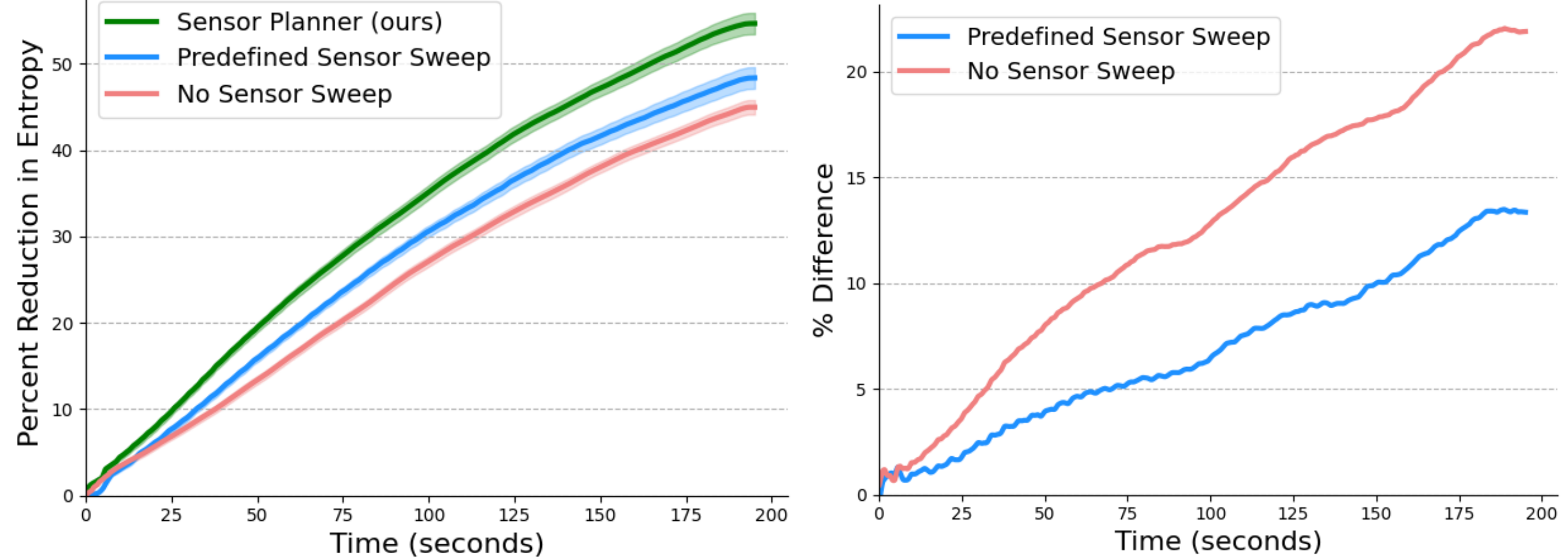}
    \caption{Results of 100 runs comparing our proposed approach to a system with a system having a predefined sensor sweep and a system having no sensor sweeping. The shaded regions represent the 95\% confidence intervals for the mean percent reduction in entropy between the approaches. The second figure shows the percentage difference between our sensor planner and the two approaches. }
    \label{fig:endToendALL}
\end{figure}

\section{Results}\label{sec:results}

To evaluate our method, we compared it against a predefined constant sensor sweep and a static no sensor sweep baseline in various scenarios. We implemented all three of these approaches in C++. We first tested the performance of our approach through a hundred runs using a simplified simulator that assumed fixed roll and pitch for the vehicle. For these runs, we initialized a uniform search map of size 5000 x 5000 meters with randomized start positions for the UAV. The UAV had a budget of $B=5000$ and a fixed camera pitch of $\theta_s = 10\degree$ with the sweep bounds of the gimbal being $\psi_{min} = - 30\degree$ and $\psi_{max} = 30\degree$. 
The results from these runs can be seen in Figure \ref{fig:endToendALL}. Our results showed an improvement of \textit{21.88\%} over the static sensor baseline and an improvement of \textit{13.34\%} over the constant sweep baseline. 



\begin{table}[h!]
\centering
\caption{Comparison of Final Percent Reduction and Rates of Percent Reduction in Entropy for two Scenarios in Isaac Sim}
\label{tab:entropy_reduction_combined}
\begin{tabularx}{\columnwidth}{@{}lXXXX@{}}
\toprule
 & \multicolumn{2}{c}{\textbf{\% Reduction}} & \multicolumn{2}{c}{\textbf{\% Reduction / second}} \\ 
\textbf{Approach} & A & B & A & B \\  \cmidrule(r){1-1}\cmidrule(lr){2-3}\cmidrule(lr){4-5}
\textbf{Sensor Planner (ours)} & \textbf{66.05} & \textbf{85.09} & \textbf{0.56} & \textbf{0.74} \\
No Sensor Sweep & 34.23 & 17.49 & 0.30  & 0.15\\
Predefined Sensor Sweep & 59.54 & 77.60 & 0.51 & 0.65 \\ \bottomrule
\end{tabularx}
\end{table}




\begin{figure*}[t]
    \centering
    \includegraphics[trim={0cm 0cm 0cm 0cm}, width=\textwidth]{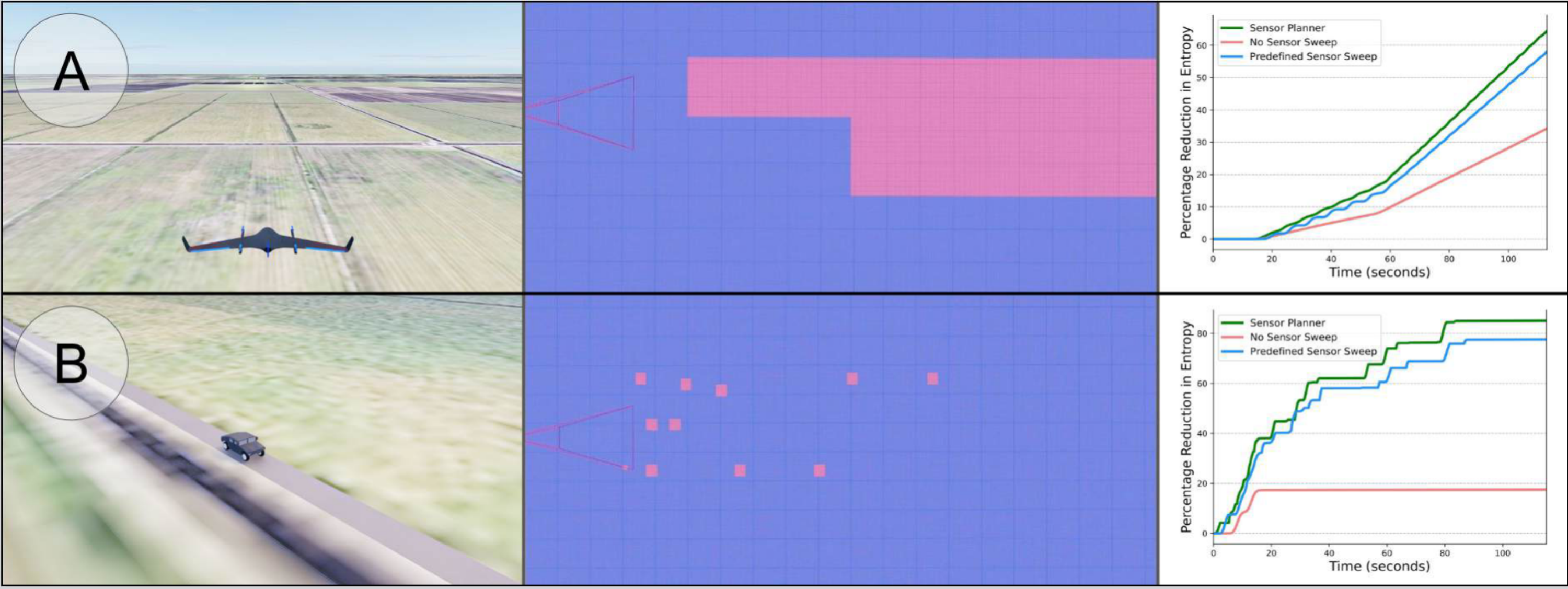}
    \caption{Scenario A and B show our approach in two different environments. In Scenario A, the UAV is given a subset of the open fields to search for a car. In Scenario B, the UAV is given distinct points where a car could be. }
    \label{fig:IsaacSim}
\end{figure*}

To test our method in an environment closer to real-world conditions, we also performed two tests in an NVIDIA Isaac Sim with Gazebo and PX4-Autopilot using the same sensor planner parameters. Isaac Sim was used to generate perception data, while Gazebo provided the dynamics of the UAV. PX4-Autopilot coupled with MAVROS was used to control the UAV. The plotted results from these tests can be seen in Fig~\ref{fig:IsaacSim}. The two environments we tested were an environment with large regions of varying priority and an environment with specific points of interest. Table \ref{tab:entropy_reduction_combined} puts forward the numerical results of the above-mentioned tests conducted in Isaac Sim.


From these results, we can see the benefits of our approach in a variety of environments. In the case of our proposed sensor planner, the percentage reduction in entropy is always greater compared to the two baseline approaches. Fig.~\ref{fig:endToendALL} shows us the mean plot of the three approaches from the 100 tests conducted in a simplified simulation environment. Having a greater percent entropy reduction and a greater percent rate of entropy reduction per second is directly proportional to a more enhanced and optimal search mission. Similarly, with the Isaac Sim tests, we observe a similar trend in the entropy reduction plots. Table \ref{tab:entropy_reduction_combined} provides us the final value of percent entropy reduction for the two testing environments in Isaac Sim. Our proposed approach again outperforms the baselines, having a 66.05\% entropy reduction with a rate of increase of 0.56 for Scenario A which has large high-priority regions. For Scenario B which involves specific points of interest, there was an 85.09\% entropy reduction and 0.74 rate of reduction. The fixed forward-facing camera ends up missing a lot of information because it's bounded by the orientation of the UAV. On the other hand, the predefined sensor sweep method wastes a lot of time looking at areas that have already been viewed or have low information gain. Our method strikes a balance between these two methods by prioritizing viewing regions of higher reward to maximize our overall information gain.

\section{Conclusion \& Future Work}\label{sec:conclusion}
This work presents a novel sensor planning framework integrated with a global IPP planner to enhance search efficiency and optimize data acquisition, thus rapidly reducing the system entropy over the belief space. The end-to-end results tested on hundreds of different search spaces demonstrate the importance of having a sensor planner in an IPP system. These results are further validated by Isaac Sim tests after running the sensor planner integrated system in two simulation environments. 

Future work could include extensive field testing to validate our simulation results.
Another extension could be to allow for controlling the full gimbal pose and fully optimizing the trajectory of the sensor. This was beyond the scope of our problem formulation and interface constraints, but it could result in smoother trajectories that effectively utilize the information from our belief space. 

\addtolength{\textheight}{-20.77cm}   







\bibliographystyle{ieeetr}
\bibliography{root}

\end{document}